%% file: main.tex
\documentclass[10pt,twocolumn,letterpaper]{article}
\usepackage[pagenumbers]{cvpr}
\input{preamble}
\definecolor{cvprblue}{rgb}{0.21,0.49,0.74}
\usepackage[pagebackref,breaklinks,colorlinks,allcolors=cvprblue]{hyperref}
\hypersetup{
  pdftitle={STAR: Counterfactual Dual-World Rollouts and Spatial Target Adaptive Reconstruction for Visual On-Policy Distillation},
  pdfauthor={Yi Zou, Zhuang Yan, Yudong Zhang, Xu Wang, Yang Wang},
  pdfsubject={Visual on-policy distillation},
  pdfkeywords={STAR, vision-language models, on-policy distillation}
}
\input{resource_links}
\date{}
\title{STAR: Counterfactual Dual-World Rollouts and Spatial Target Adaptive Reconstruction for Visual On-Policy Distillation}
\makeatletter
\renewcommand{\@maketitle}{%
  \newpage\null\vskip 15pt
  \begin{center}
    {\Large\bfseries\@title\par}
    \vspace*{3pt}
    {\lineskip 1pt\begin{tabular}[t]{@{}c@{}}\@author\end{tabular}\par}
    \vspace*{1pt}
  \end{center}
}
\makeatother
\author{%
{\fontsize{11}{12}\selectfont
Yi Zou$^{1,*}$\quad Zhuang Yan$^{2}$\quad Yudong Zhang$^{1}$\quad
Xu Wang$^{1,\ddagger}$\quad Yang Wang$^{1,\ddagger}$}\\[-1pt]
{\fontsize{8.5}{9.5}\selectfont
$^{1}$University of Science and Technology of China, Hefei, China\qquad
$^{2}$AMAP, Alibaba Group}\\[-1pt]
{\fontsize{8}{9}\selectfont\texttt{yizou@mail.ustc.edu.cn}\quad
\texttt{yanzhuang.yz@alibaba-inc.com}\quad
\texttt{\{yudong.zhang,wx309,angyan\}@ustc.edu.cn}}\\[1pt]
\STARResources
}

\begin{document}
\maketitle
\begingroup
\renewcommand{\thefootnote}{\fnsymbol{footnote}}
\footnotetext[1]{Work done during internship at AMAP, Alibaba Group.}
\footnotetext[3]{Corresponding authors.}
\endgroup
\input{sec/0_abstract}
\input{sec/1_intro}
\input{sec/2_related_work}
\input{sec/3_method}
\input{sec/4_experiments}

\FloatBarrier
\input{sec/5_conclusion}
\clearpage
\input{sec/llm_usage}
{
    \small

\input{main.bbl}
}
\end{document}

%% file: preamble.tex
\usepackage{microtype}
\usepackage{array,colortbl,tabularx}
\usepackage{tikz}
\usepackage{placeins}
\definecolor{takeawayblue}{RGB}{35,79,126}
\DeclareRobustCommand{\resultlock}{%
  \begingroup\setlength{\unitlength}{1pt}\color{takeawayblue}%
  \begin{picture}(7,8)
    \linethickness{0.7pt}%
    \put(3.5,4.8){\oval(4,5.5)[t]}%
    \put(0.5,0){\rule{6pt}{5pt}}%
    \put(3.5,3.1){\color{white}\circle*{1.2}}%
    \put(3.2,1.6){\color{white}\rule{0.6pt}{1.4pt}}%
  \end{picture}\endgroup}
\newcommand{\takeaway}[2]{%
  \par\addvspace{3pt}%
  \begingroup
  \setlength{\parskip}{0pt}%
  \noindent\begin{tikzpicture}
    \node[draw=black!90,fill=black!4,rounded corners=4pt,
      line width=0.6pt,inner xsep=5pt,inner ysep=4pt,outer sep=0pt,
      text width=\dimexpr\linewidth-10.6pt\relax,
      align=justify,font=\normalfont\itshape,text=black]
      {\textbf{Takeaway #1.}~#2};
  \end{tikzpicture}%
  \par\endgroup\addvspace{3pt}%
}

\AtBeginDocument{%
  \flushbottom
  \setlength{\abovedisplayskip}{6pt plus 1pt minus 1pt}%
  \setlength{\belowdisplayskip}{6pt plus 1pt minus 1pt}%
  \setlength{\abovedisplayshortskip}{0pt plus 1pt}%
  \setlength{\belowdisplayshortskip}{4pt plus 1pt minus 1pt}%
}

%% file: resource_links.tex
\newcommand{\STARCodeURL}{https://github.com/xiaolingdudu/STAR}
\newcommand{\STARDatasetURL}{https://huggingface.co/collections/Yizou/star}
\newcommand{\STARModelURL}{https://huggingface.co/collections/Yizou/star}

\newcommand{\STARResources}{%
  {\fontsize{9}{10}\selectfont
  \href{\STARCodeURL}{%
    \raisebox{-2pt}{\includegraphics[height=9pt]{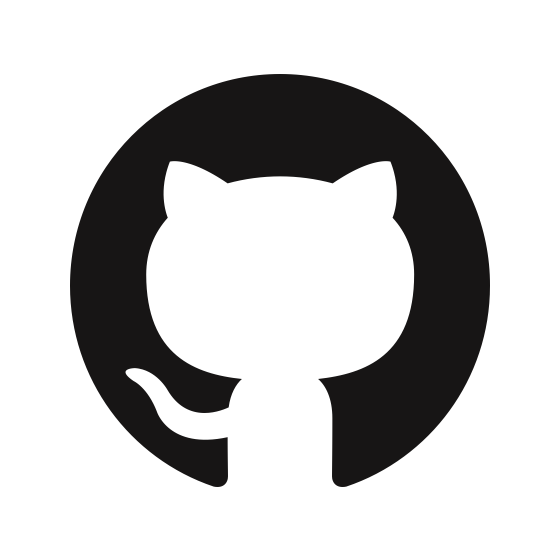}}\enspace Code}%
  \qquad
  \href{\STARDatasetURL}{%
    \raisebox{-2pt}{\includegraphics[height=10pt]{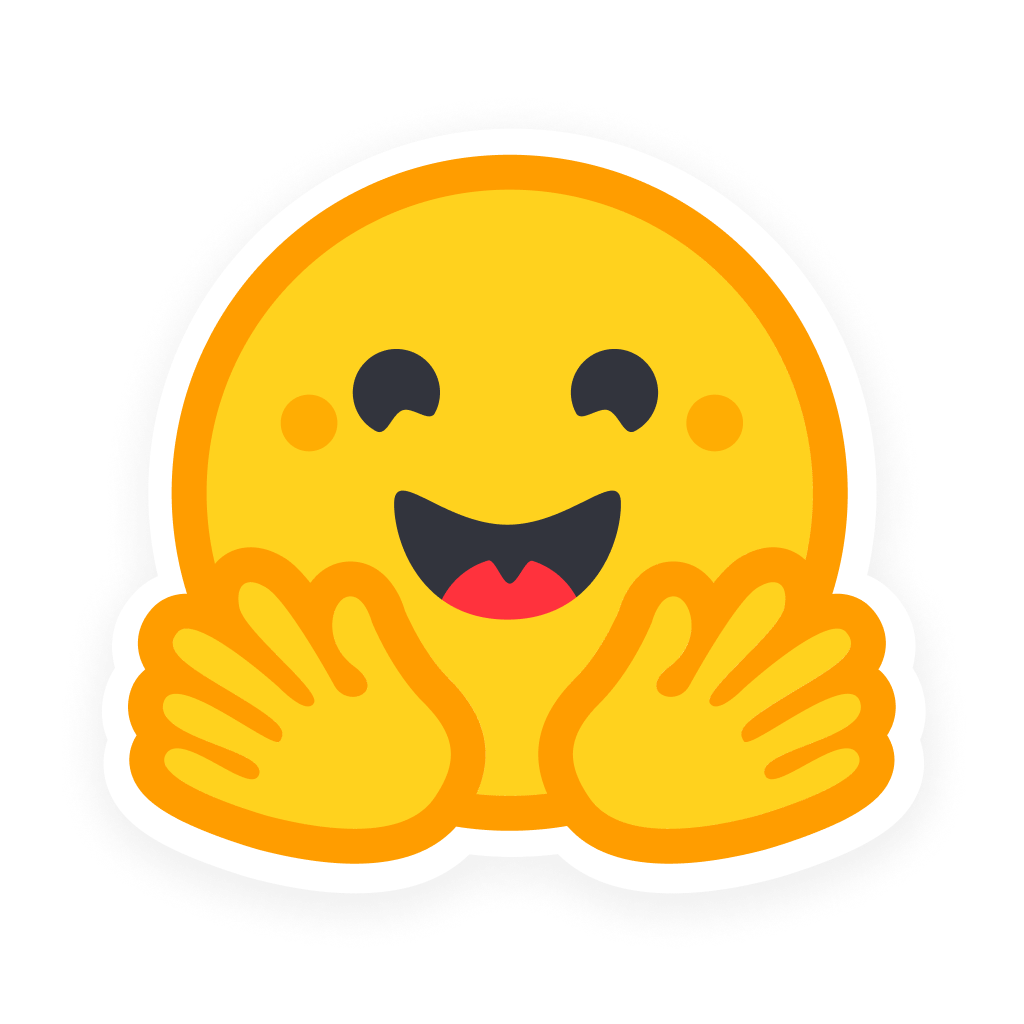}}\enspace Dataset}%
  \qquad
  \href{\STARModelURL}{%
    \raisebox{-2pt}{\includegraphics[height=10pt]{fig/resource-icons/huggingface.png}}\enspace Model}%
  }%
}

%% file: sec/0_abstract.tex
\begin{abstract}
Vision-language models often rely on global context and language priors while overlooking answer-determining local details.
Privileged visual on-policy self-distillation supervises full-image student rollouts using a frozen teacher that views high-resolution evidence crops.
Yet our cross-scale diagnostics show that models retain over half of their full-image accuracy when local evidence is completely masked, indicating that single-image correctness does not guarantee reliance on local evidence, while the need for visual correction also varies across tokens with their spatial attention states.
We propose \textbf{STAR}, a visual on-policy self-distillation framework combining independent counterfactual dual-world rollouts with spatial-attention-guided target reconstruction.
Local image editing constructs paired worlds with identical questions and backgrounds but distinct answer-determining facts, where a shared student samples trajectories independently to expose contextual shortcuts under image-consistent prefixes.
Within each world, the teacher contrasts real and mean-color crops to determine the visual correction direction, while the divergence between student image attention and a teacher-centroid spatial reference adaptively controls token-wise target log-probability reconstruction.
Without auxiliary attention losses or test-time tool overhead, STAR improves Qwen3.5 across scales by up to 8.04 percentage points in average accuracy over initialization across six fine-grained benchmarks, reaches OPSD's final accuracy $4.7\times$ faster, and generalizes to out-of-domain reasoning tasks.
\end{abstract}

%% file: sec/1_intro.tex
\section{Introduction}
\label{sec:intro}

Vision-language models (VLMs) have achieved remarkable progress in general visual understanding and reasoning, yet they frequently overlook decisive local details in complex or high-resolution scenes, defaulting instead to language priors and global contextual cues~\cite{tong2024cambrian,tong2024eyes}.
While supervised fine-tuning can instill visual knowledge through reference trajectories, it requires costly expert demonstrations and suffers from exposure bias caused by the mismatch between fixed training prefixes and self-generated prefixes at inference~\cite{agarwal2024onpolicydistillationlanguagemodels,liu2023visualinstructiontuning}.
Reinforcement learning (RL) directly optimizes self-generated rollouts on-policy, but hinges on verifiable outcome rewards and incurs heavy exploration overhead~\cite{liu2025visualrft,zheng2026deepeyes}.
Recently, privileged visual on-policy self-distillation has emerged as a compelling alternative: a full-image student generates on-policy rollouts, while a frozen teacher initialized from the same checkpoint views a high-resolution local evidence crop and supplies dense, token-wise distributional supervision along the student's own prefixes~\cite{yuan2026visionopdlearningfinedetails}.
This paradigm provides dense supervision at generation states actually visited by the student, mitigating exposure bias and reducing reliance on stronger external teachers for fine-grained multimodal post-training.

\begin{figure*}[t]
    \centering
    \includegraphics[width=\textwidth]{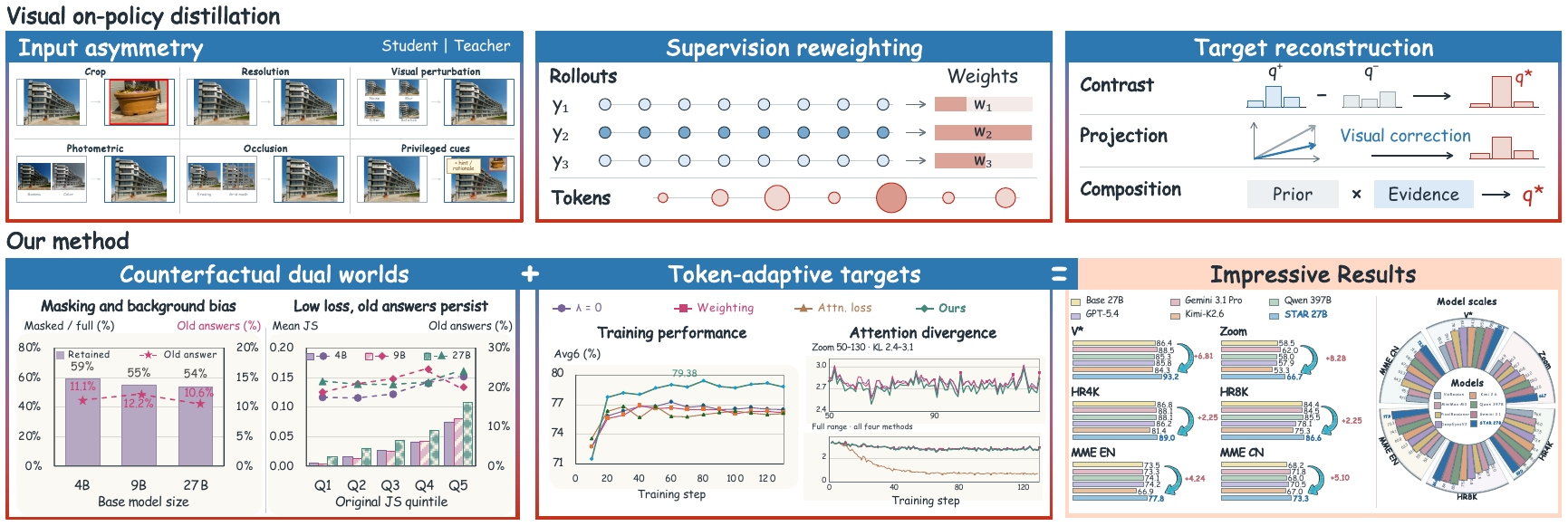}
   \caption{Overview of existing visual on-policy distillation paradigms and the motivation of STAR. The top row summarizes prior interventions on input conditions, supervision weights, and target distributions. In the bottom row, left cross-scale masking diagnostics expose contextual shortcuts and old-answer persistence in single-world training; middle 4B training dynamics show that adaptive target reconstruction achieves higher Avg6 than raw teacher targets, external loss weighting, and explicit attention loss; and right benchmark comparisons show that STAR improves over base models while surpassing leading closed-source, open-weight, and visual tool models.}
    \label{fig:teaser}
\end{figure*}

As illustrated in the top row of Figure~\ref{fig:teaser}, existing studies primarily intervene at three levels: input conditions, supervision weights, and target distributions.
Privileged context, local crops, resolution differences, and perturbed views create information asymmetry between student and teacher, giving the teacher richer visual or semantic information~\cite{shenfeld2026selfdistillationenablescontinuallearning,tian2026vicurvisualcuesrecoverable,wang2026selfboostingvisionlanguagemodelsnoisy,yuan2026visionopdlearningfinedetails,zhu2026rpopsdresolutionprivilegedonpolicyselfdistillation}.
Weighting and filtering based on visual evidence relevance concentrate supervision on informative response trajectories or tokens~\cite{aniri2026opdvvisualonpolicyselfdistillation,hu2026evidencealignedmultimodalonpolicyselfdistillation,liu2026visualadvantageonpolicydistillationvisionlanguage,sun2026vzeroanswerlabelfreeonpolicydistillation}.
Target reconstruction and optimization constraints further modify relative candidate probabilities or update directions to strengthen the influence of visual evidence during learning~\cite{liang2026visualcontrastiveselfdistillation,shin2026legoopdfactorizedteachercomposition,wang2026cvopdcontrastivevisiononpolicy,zhang2026vadattributingvisualevidence}.
Although these advances improve supervision quality, reliable reliance on local evidence also depends on whether training rollouts expose changes in local facts and whether token-wise supervision matches the visual attention state at the current generation position.

Despite increasingly rich visual supervision, correct student responses on original images may still rely excessively on background correlations and language priors.
We study 6,176 image pairs that preserve the question and background while editing only answer-determining local facts.
As shown in the lower-left panel of Figure~\ref{fig:teaser}, masking the evidence region leaves the 4B, 9B, and 27B models with 59.1\%, 54.9\%, and 53.9\% of their full-image accuracy, respectively.
Even on pairs where the original full image and both evidence-only views on mean-color canvases are answered correctly, the models still choose the original answer for 10.6\% to 12.2\% of the edited full images.
Such contextual dependence is further concealed by teacher--student agreement in a single world: among the 20\% of original-correct trajectories with the lowest JS divergence between teacher and student predictions, the old-answer rates on edited images remain 17.45\%, 18.73\%, and 21.52\% for 4B, 9B, and 27B, respectively.
These empirical results demonstrate that single-image correctness does not guarantee genuine reliance on local visual evidence, motivating independent dual-world rollouts where the student generates responses conditioned on each world's distinct visual facts.

\takeaway{1}{Correct single-world responses often reflect background shortcuts; counterfactual dual worlds expose and counteract insensitivity to local visual changes.}

Learning from local fact changes also requires visual correction suited to each generation position.
Attention varies across positions within a response, and erroneous prefixes can weaken the teacher's preference for correct visual candidates~\cite{qiu2026opdahalinguisticmomentumvisual}.
Positional weighting scales a token's gradient contribution without changing relative candidate probabilities within a fixed target.
Visual contrast modifies these probabilities, but a uniform correction magnitude may not suit all positions.
Prior methods mainly derive token-wise corrections from vocabulary output distributions~\cite{shin2026legoopdfactorizedteachercomposition,zhang2026vadattributingvisualevidence}.
Student spatial attention offers another direct basis for matching reconstruction strength to attention over local evidence, amplifying visual contrast when focus drifts while keeping well-grounded positions smoothly regularized.
As shown in the middle panel of Figure~\ref{fig:teaser}, adaptive target reconstruction achieves the highest Avg6 from step~20 onward, whereas directly minimizing attention divergence lowers online attention KL without yielding the highest response accuracy.

\takeaway{2}{Visual correction strength should adapt to each token's spatial attention to local evidence.}

Motivated by these findings, we propose \textbf{STAR}, a visual on-policy self-distillation framework that combines independent counterfactual dual-world rollouts with \textbf{S}patial \textbf{T}arget \textbf{A}daptive \textbf{R}econstruction guided by spatial attention.
Through local image editing, we construct two visual worlds with the same background and question but different correct answers.
A shared student independently generates responses in both worlds, with every trajectory supervised under its corresponding image condition.
Local fact changes thus enter the generation states actually visited by the student, while the reasoning content and response lengths can evolve naturally with the visual facts in each world.
Within each world, the teacher's prediction difference between the real evidence crop and a mean-color crop provides a visual correction direction.
We construct a spatial reference from the positive teacher's local attention centroid and use the divergence between the student's image attention and this reference to modulate the reconstruction strength of teacher log probabilities at each token.
Together, these mechanisms incorporate local fact changes and token-wise evidence attention states into training.
At inference, the student generates responses directly from the full image and question, without additional bounding boxes or a teacher branch, while preserving the standard inference pipeline.

Our main contributions are:
\begin{itemize}
    \item Cross-scale diagnostics show that single-world accuracy conceals background shortcuts and uniform correction overlooks token-wise attention to local evidence.
    \item We propose STAR, combining independent counterfactual dual-world on-policy rollouts with spatial-attention-guided token-wise teacher target reconstruction.
    \item Experiments across eight models spanning diverse scales and dense and MoE architectures show consistent gains over base models, tool agents, and frontier VLMs.
\end{itemize}

%% file: sec/2_related_work.tex
\section{Related Work}
\label{sec:related_work}

\begingroup
\setlength{\parskip}{0pt}
\noindent\textbf{Visual On-Policy Distillation.}
Conventional offline SFT updates autoregressive models on fixed reference trajectories, creating a training–inference state mismatch that compounds errors whenever generation deviates from the reference~\cite{ross2011reduction,bengio2015scheduled,agarwal2024onpolicydistillationlanguagemodels}.
On-policy distillation addresses this exposure bias by obtaining teacher next-token supervision along student-sampled trajectories~\cite{agarwal2024onpolicydistillationlanguagemodels}.
Supplying the teacher with privileged context further enables self-distillation from an identically initialized checkpoint without relying on a stronger external teacher~\cite{shenfeld2026selfdistillationenablescontinuallearning},
inspiring recent analyses into its underlying mechanisms and contextual reliability~\cite{hou2026uniopdunifyingonpolicydistillation,li2026rethinkingonpolicydistillationlarge,ding2026doesonpolicydistillationreally}.
This paradigm has been extended to vision-language models by pairing a full-image student with a high-resolution evidence-crop teacher that provides fine-grained supervision along student-generated prefixes~\cite{yuan2026visionopdlearningfinedetails}.
Related approaches have also shown promising results in temporal video grounding~\cite{li2026videoopdefficientposttrainingmultimodal} and GUI grounding~\cite{zhang2026learnclickyourselfonpolicy}.
Visual input asymmetry can be constructed through cues recoverable at test time~\cite{tian2026vicurvisualcuesrecoverable}, resolution differences~\cite{zhu2026rpopsdresolutionprivilegedonpolicyselfdistillation}, and noisy or augmented student views~\cite{wang2026selfboostingvisionlanguagemodelsnoisy,li2026selfsupervisedvisualonpolicydistillation}.
Crucially, however, both single-image rollouts and semantics-preserving augmentations leave the answer-determining local facts invariant, allowing spurious background–answer correlations to survive training. By contrast, we construct counterfactual dual worlds that preserve the background and question while altering decisive local facts and ground-truth answers, sampling on-policy trajectories independently in each world to break contextual shortcuts across student-visited states.

\par\addvspace{6pt}
\noindent\textbf{Refining Distillation Supervision.}
Even with privileged visual inputs, self-distillation remains vulnerable to language priors and contextual bias. In long-form chain-of-thought generation, most tokens act as linguistic scaffolding while only a sparse subset governs visual facts~\cite{liu2026visualadvantageonpolicydistillationvisionlanguage}.
Consequently, raw teacher distributions can be compromised by tension between linguistic imitation and visual grounding~\cite{yoon2026decomposedonpolicydistillationvisionlanguage}, spurious teacher signals~\cite{jiang2026teachersmisleadspurioussignalawareonpolicy}, and erroneous student prefixes that bias the teacher's preference among visual candidates during subsequent autoregressive response generation~\cite{qiu2026opdahalinguisticmomentumvisual}.
To isolate visual supervision, recent methods contrast clear inputs against degraded, masked, or erased views along two main lines. The first line applies external evidence-aware weighting or selective filtering to prioritize visually informative trajectories and tokens~\cite{sun2026vzeroanswerlabelfreeonpolicydistillation,liu2026visualadvantageonpolicydistillationvisionlanguage,aniri2026opdvvisualonpolicyselfdistillation,hu2026evidencealignedmultimodalonpolicyselfdistillation,xie2026reinforcementlearningvisuallyattributable,ye2026editdontimitateminimaledit}.
however, scalar loss weights only rescale gradients and cannot repair visual distinctions already diluted within a fixed teacher target. The second line alters optimization constraints or directly reconstructs the teacher’s target distribution in vocabulary space~\cite{wang2026cvopdcontrastivevisiononpolicy,zhang2026vadattributingvisualevidence,liang2026visualcontrastiveselfdistillation,qiu2026opdahalinguisticmomentumvisual}.
Extending this latter direction, STAR couples real-versus-mean-color crop log-probability contrast with student–teacher spatial attention divergence, dynamically calibrating the target reconstruction strength at each token according to whether the student is attending to the relevant local evidence.
\par\endgroup

%% file: sec/3_method.tex
\section{Method}
\label{sec:method}

\begin{figure*}[t]
    \centering
    \includegraphics[width=\textwidth]{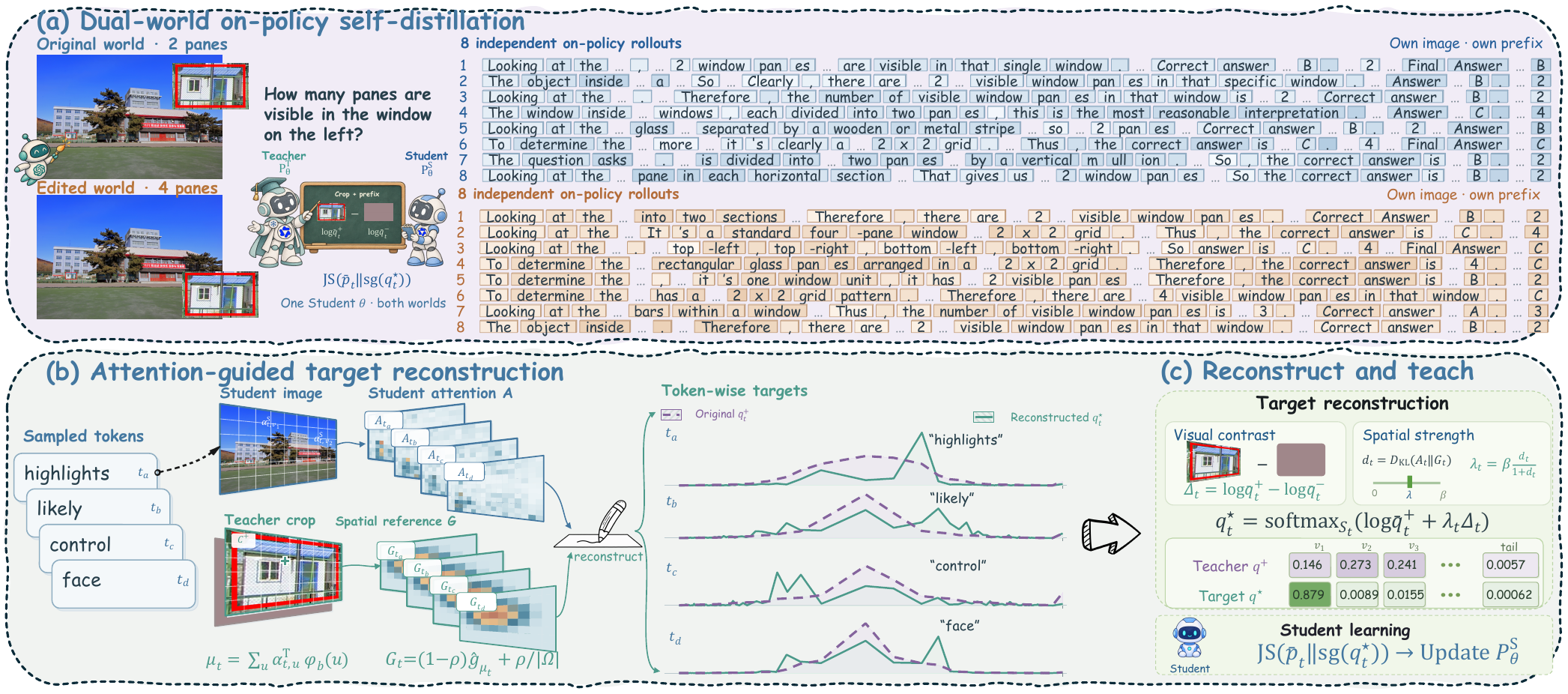}
           \caption{Overview of STAR. (a) A shared student independently samples on-policy rollouts in original and counterfactually edited worlds, while a frozen teacher evaluates each rollout conditioned on its own prefix. (b) Student attention is compared with a spatial reference constructed from teacher crop attention to calibrate the reconstruction strength at each sampled token position. (c) Real-versus-mean-color crop contrast and the calibrated strength jointly reconstruct the target distribution, which supervises the student through Jensen--Shannon distillation. Purple dashed and green solid curves denote the original teacher and reconstructed target probability distributions, respectively.}

    \label{fig:overview}
\end{figure*}

Figure~\ref{fig:overview} illustrates the overall architecture of STAR.
We first formulate visual on-policy distillation under a unified framework (Section~\ref{sec:preliminaries}),and then detail our two core mechanisms: counterfactual dual-world on-policy rollouts (Section~\ref{sec:dual_world}) and spatial-attention-guided token-wise target reconstruction (Section~\ref{sec:target_recon}), before presenting the overall training objective (Section~\ref{sec:distill_objective}) for learning jointly across the paired worlds.

\subsection{Preliminaries and Unified Formulation}
\label{sec:preliminaries}

Given a question $x$, a full image $I$, and a local high-resolution evidence crop $C$, let $P_\theta^S$ denote the trainable student and $P_\theta^T$ the frozen teacher initialized from the same pretrained checkpoint.
Only the student is updated during training.
At each step, the student autoregressively samples a response $\hat y=(\hat y_1,\dots,\hat y_T)\sim P_\theta^S(\cdot\mid I,x)$.
Conditioned on the generated prefix $h_t=\hat y_{<t}$, the student and teacher next-token distributions over vocabulary $\mathcal V$ are
\begin{equation}
    \begin{aligned}
        p_t(z)&=P_\theta^S(z\mid I,x,h_t),\\
        q_t^+(z)&=P_\theta^T(z\mid C,x,h_t),
        \qquad z\in\mathcal V.
    \end{aligned}
    \label{eq:student_teacher_distributions}
\end{equation}

The general objective of visual on-policy distillation is
\begin{equation}
    \begin{aligned}
        \mathcal L_{\mathrm{gen}}(\theta)
        &=\mathbb E_{(x,I,C)\sim\mathcal W}\,
        \mathbb E_{\hat y\sim P_\theta^S(\cdot\mid I,x)}\\
        &\quad\left[
            \frac{1}{|\hat y|}
            \sum_{t=1}^{|\hat y|}
            \omega_t\,
            D\!\left(p_t\,\middle\|\,\operatorname{sg}(\tilde q_t)\right)
        \right],
    \end{aligned}
    \label{eq:unified_opd}
\end{equation}
where $\mathcal W$ samples training examples and visual conditions, $\omega_t\geq0$ is a scalar token weight, $\tilde q_t$ is the target distribution, $\operatorname{sg}(\cdot)$ stops gradients, and $D(\cdot\Vert\cdot)$ is a distributional divergence.
The standard setting~\cite{yuan2026visionopdlearningfinedetails} uses only the original image per example, with $\omega_t=1$ and $\tilde q_t=q_t^+$.
With $\tilde q_t$ fixed, varying $\omega_t$ only scales the gradient contribution at position $t$ without changing the relative candidate probabilities within the fixed target distribution $\tilde q_t$ at that token.

In Equation~\eqref{eq:unified_opd}, $\mathcal W$ and the student policy jointly determine the visual conditions and generation states receiving supervision, while $\tilde q_t$ specifies the relative candidate probabilities learned in each state.
Keeping external weights uniform ($\omega_t=1$), we modify both components: we construct counterfactual dual worlds with altered local facts and sample independently in each, and we adaptively reconstruct the teacher target according to the student--teacher visual attention divergence at the current response position.

\subsection{Counterfactual Dual-World On-Policy Rollouts}
\label{sec:dual_world}

We construct counterfactual dual worlds from Vision-OPD~\cite{yuan2026visionopdlearningfinedetails}, with similar backgrounds but different decisive local facts.
For instance $i$, let $x_i$ denote the question and answer options, $b_i$ the annotated evidence bounding box, and $I_i^{(0)}$ and $C_i^{(0)}$ the original full image and its corresponding evidence crop.
Based on the question and options, we select a visually realizable target option different from the original answer and formulate an instruction to edit the answer-determining local content.
Using the original crop as a reference, an image editing model generates the edited crop $C_i^{(1)}$ under instructions to preserve the original composition, background, and unrelated details.
We resize the generated crop to match the original crop and paste it into the corresponding location in the full image to obtain $I_i^{(1)}$, leaving content outside the pasted region unchanged.
Both worlds share the same question and options but have distinct correct answers $a_i^{(0)}\neq a_i^{(1)}$.
We use $w\in\{0,1\}$ to index the paired original and edited worlds, respectively.

Both visual worlds share a single student model $P_\theta^S$.
For each paired instance, we independently sample $K$ on-policy responses conditioned on each world $w\in\{0,1\}$:
\begin{equation}
    \hat y_{i,k}^{(w)}\sim P_\theta^S\!\left(\cdot\mid I_i^{(w)},x_i\right),
    \quad k=1,\ldots,K.
    \label{eq:dual_world_rollouts}
\end{equation}
These sampled discrete responses are held fixed during the parameter update.
Each world generates and uses its own response trajectories, allowing the reasoning content and sequence lengths to differ.
Omitting the instance index $i$ and rollout index $k$, we denote the prefix before the $t$-th response position in world $w$ by $h_{w,t}=\hat y_{<t}^{(w)}$.
The corresponding student distribution is
\begin{equation}
    p_{w,t}(z)=P_\theta^S\!\left(z\mid I^{(w)},x,h_{w,t}\right),
    \quad z\in\mathcal V.
    \label{eq:world_student_distribution}
\end{equation}
The frozen teacher uses the evidence crop from the corresponding world to provide supervision along that world's student-generated prefixes, keeping the image conditions consistent with the trajectories and covering the generation states actually visited by the student policy.

\subsection{Attention-Guided Target Reconstruction}
\label{sec:target_recon}

Within each world $w$, the frozen teacher $P_\theta^T$ evaluates the same student prefix $h_{w,t}$ under the real evidence crop $C^{(w)}$ and a same-sized mean-color crop $\widetilde C^{(w)}$, yielding the distributions $q_{w,t}^+$ and $q_{w,t}^-$, respectively.
The mean-color crop is filled with the average RGB value of $C^{(w)}$ to remove local spatial details.
We combine the student's and positive teacher's respective top-$k$ candidates and aggregate each distribution's remaining vocabulary probability mass into a single $\mathrm{tail}$ entry, forming a shared support $\mathcal S_{w,t}$.
The resulting compressed distributions over the shared support $\mathcal S_{w,t}$ are denoted by $\bar p_{w,t}$, $\bar q_{w,t}^+$, and $\bar q_{w,t}^-$, respectively.

We use spatial attention disagreement to determine the reconstruction strength at each response position.
At a selected attention layer, we apply a softmax over visual positions within each attention head and average across heads to obtain the student's image attention distribution $A_{w,t}$ over the full-image grid $\Omega_w$.
The positive teacher's crop attention $A_{w,t}^T$ is extracted in the same way at the corresponding layer and prediction position.
We map its attention centroid to full-image coordinates and center a two-dimensional Gaussian $g_{w,t}$ at this location, with spatial scales determined by the evidence box $b$.
Mixing the normalized Gaussian with a uniform distribution of weight $\rho$ gives the spatial reference
\begin{equation}
    G_{w,t}(v)
    =(1-\rho)\frac{g_{w,t}(v)}{\sum_{u\in\Omega_w}g_{w,t}(u)}
    +\frac{\rho}{|\Omega_w|},
    \label{eq:teacher_spatial_reference}
\end{equation}
for each visual position $v\in\Omega_w$.
We measure the difference between the student's image attention and this reference using KL divergence and map it to a bounded correction coefficient:
\begin{equation}
    d_{w,t}=D_{\mathrm{KL}}\!\left(A_{w,t}\,\middle\|\,G_{w,t}\right),
    \quad \lambda_{w,t}=\beta\frac{d_{w,t}}{1+d_{w,t}}.
    \label{eq:attention_reconstruction_strength}
\end{equation}
The coefficient remains small when $A_{w,t}$ is close to $G_{w,t}$ and increases monotonically with their divergence, with $\beta$ controlling the maximum visual correction strength.

For each candidate $z\in S_{w,t}$, the log-probability contrast
$\Delta_{w,t}(z)=\log\bar q_{w,t}^{+}(z)-\log\bar q_{w,t}^{-}(z)$
captures the net evidence gain contributed by fine-grained local details over the mean-color baseline.
Combining this visual contrast direction with the attention-modulated coefficient $\lambda_{w,t}$, we reconstruct the target distribution on $S_{w,t}$ as
\begin{equation}
    \log q_{w,t}^{\star}(z)
    =
    \log\bar q_{w,t}^{+}(z)
    + \lambda_{w,t}\Delta_{w,t}(z)
    - \log Z_{w,t},
    \label{eq:target_reconstruction}
\end{equation}
where $Z_{w,t}=\sum_{u\in S_{w,t}}\bar q_{w,t}^{+}(u)
\exp\!\bigl(\lambda_{w,t}\Delta_{w,t}(u)\bigr)$
is the partition function.
Intuitively, $\bar q_{w,t}^{+}$ provides a fluent base distribution,
$\Delta_{w,t}$ selectively boosts visually grounded candidates over prior-driven ones,
and $\lambda_{w,t}$ dynamically gates this visual amplification according to the student's real-time spatial attention state.

\subsection{Distillation Objective}
\label{sec:distill_objective}

During optimization, we stop gradients through the reconstructed target, denoting it by $\widetilde q_{w,t}=\operatorname{sg}(q_{w,t}^\star)$, and form the midpoint distribution $m_{w,t}=\frac{1}{2}(\bar p_{w,t}+\widetilde q_{w,t})$ on the shared support $\mathcal S_{w,t}$.
The distillation loss at each valid response position is the Jensen--Shannon divergence on this support:
\begin{equation}
    \ell_{w,t}
    =\frac{1}{2}D_{\mathrm{KL}}\!\left(\bar p_{w,t}\,\middle\|\,m_{w,t}\right)
    +\frac{1}{2}D_{\mathrm{KL}}\!\left(\widetilde q_{w,t}\,\middle\|\,m_{w,t}\right).
    \label{eq:distillation_js}
\end{equation}
We aggregate these losses over valid response positions from both worlds to train a single student.
The spatial divergence $d_{w,t}$ is used only to compute the detached reconstruction coefficient and does not introduce an auxiliary attention loss.
Gradients propagate only through the student predictions $\bar p_{w,t}$.
At inference time, the trained student $P_\theta^S$ answers questions directly from the full image.

%% file: sec/4_experiments.tex
\section{Experiments}
\label{sec:experiments}

\input{tab/main_results}

\subsection{Experimental Settings}
\label{sec:exp_settings}

\begingroup
\setlength{\parskip}{0pt}
\noindent\textbf{Datasets.}
Following~\cite{sun2026vzeroanswerlabelfreeonpolicydistillation,yuan2026visionopdlearningfinedetails}, we evaluate the ability to localize, recognize, and reason about fine visual details across six high resolution benchmarks: V* Bench~\cite{wu2023vstar}, ZoomBench~\cite{wei2026zooming}, HR-Bench (4K and 8K)~\cite{wang2024divide}, and MME-RealWorld (English and Chinese)~\cite{zhang2024mmerealworld}.
To assess generalization beyond the training domain, we further test on DynaMath~\cite{zou2025dynamath}, MMVP~\cite{tong2024eyes}, LogicVista~\cite{xiao2024logicvista}, and CV-Bench~\cite{tong2024cambrian}, which probe visual mathematical reasoning, discrimination of fine visual details, logical reasoning, and general perception, respectively.
Our training corpus builds upon the questions, images, and bounding boxes from Vision-OPD~\cite{yuan2026visionopdlearningfinedetails}.
Using GPT-Image-2, we edit each local evidence crop toward a distractor option or alternative answer and paste it back into the original full image to form candidate pairs of original and counterfactual images.
We then use \texttt{gpt-6-astra} to verify each candidate pair, checking that the answer changes unambiguously, the edit remains confined to the evidence region, and the surrounding background is preserved.
This process yields 6,176 verified pairs, each containing matched full images and high resolution evidence crops for both worlds with different ground truth answers.

\par\noindent\textbf{Baselines.}
We benchmark STAR against three categories of competitors:
\emph{(i) Frontier VLMs with closed or open weights}, including Gemini 3 Flash, Gemini 3.1 Pro, GPT-5.2, GPT-5.4, Qwen3-VL-235B-Instruct, Qwen3.5-35B-A3B, Qwen3.5-397B-A17B, GLM-4.6V, Kimi-K2.6, Step-3.7-Flash, MiniMax-M3, and GLM-5.3-Flash;
\emph{(ii) Visual tool agents that think with images}, including DeepEyes-7B~\cite{zheng2026deepeyes}, Thyme-RL-7B~\cite{zhang2025thyme}, DeepEyesV2-7B~\cite{hong2026deepeyesv2}, PixelReasoner-7B~\cite{wang2025pixelreasoner}, VisReason-7B~\cite{li2026visreason}, and SenseNova-MARS-8B~\cite{chng2025sensenovamars}, evaluated with their native pipelines for multiple rounds of cropping and zooming on the test image;
and \emph{(iii) Posttraining methods with matched backbones} on Qwen3.5-4B and Qwen3.5-9B, spanning RL with outcome rewards (GRPO~\cite{shao2024deepseekmath}), self distillation with text hints (OPSD~\cite{zhao2026selfdistilled}), and visual on policy distillation (Vision-OPD~\cite{yuan2026visionopdlearningfinedetails}, OPD-V~\cite{aniri2026opdvvisualonpolicyselfdistillation}, VA-OPD~\cite{liu2026visualadvantageonpolicydistillationvisionlanguage}, V-Zero~\cite{sun2026vzeroanswerlabelfreeonpolicydistillation}, CV-OPD~\cite{wang2026cvopdcontrastivevisiononpolicy}, and VAD~\cite{zhang2026vadattributingvisualevidence}).
We further evaluate scalability across six additional dense and MoE backbones in Section~\ref{sec:main_results}.

\par\noindent\textbf{Metrics.} We report single-generation answer accuracy on all six visual question answering benchmarks.
Following the corresponding evaluation templates~\cite{wu2023vstar,wei2026zooming,wang2024divide,zhang2024mmerealworld}, we deploy GPT-OSS-120B through vLLM~\cite{kwon2023efficient} to judge final-answer correctness and report the equally weighted average accuracy across the six benchmarks, denoted by Avg6.
The four out-of-domain generalization benchmarks are evaluated under their standard protocols and reported separately.

\par\noindent\textbf{Implementation Details.} Both models share the Qwen3.5 initialization; training uses verl~\cite{sheng2024hybridflow} with vLLM rollouts.
Each step samples 96 image pairs and eight responses per image (1,536 trajectories), with temperature 1 and a 1,024-token limit.
We train for 130 steps on 64 NVIDIA H20 141\,GB GPUs across eight nodes using AdamW, a peak learning rate of $2\times10^{-6}$, ten-step warmup, and a constant schedule, saving checkpoints every ten steps.
We use layer-20 attention, Gaussian scales of one quarter of the box dimensions, $\rho=0.1$, and reconstruction cap $\beta=1$.
test-time inference requires only the raw evaluation image and question.
Evaluation uses non-thinking mode with temperature 0, random seed 42, and an 8,192-token limit.
\par\endgroup

\subsection{Main Results}
\label{sec:main_results}

\noindent\textbf{Main benchmark comparisons.}
Table~\ref{tab:main_results} summarizes performance on the six fine-grained visual benchmarks. STAR consistently outperforms the corresponding Qwen3.5 Base on every benchmark at both 4B and 9B, reaching Avg6 scores of 79.31\% and 78.91\%, with gains of 8.04 and 4.90 percentage points over initialization. The same-backbone comparisons support both core designs. At 4B, Vision-OPD, OPD-V, VA-OPD, and CV-OPD trained on single-world original images all lose accuracy on HR-Bench 4K and 8K, whereas counterfactual dual-world sampling exposes local fact changes and enables STAR to achieve positive gains on every benchmark at both scales. With spatial attention-guided target reconstruction, STAR-4B achieves the highest Avg6 in its group at 79.31\% and reaches 94.76\% on V* Bench, exceeding GRPO, OPSD, Vision-OPD, OPD-V, and VAD by 4.36, 4.39, 3.38, 2.02, and 0.72 percentage points in Avg6, respectively. At 9B, STAR exceeds Vision-OPD, OPD-V, CV-OPD, V-Zero, and VA-OPD by 2.81, 2.55, 1.78, 1.37, and 0.63 points in Avg6, respectively. It also exceeds VAD by 0.12 points on ZoomBench and
2.43 points on MME-RealWorld CN, achieving the highest
Chinese accuracy in the 9B group at 72.50\%. Compared with existing models, STAR-4B uses a single autoregressive generation without tool overhead to exceed the top image-tool agent SenseNova-MARS-8B by 4.01 points in Avg6, as well as Gemini 3.1 Pro at 78.04\% and open-weight Qwen3.5-397B-A17B at 76.50\%.

\input{tab/scaling}

\par\addvspace{3pt}
\par\noindent\textbf{Models scale and architecture.}
Table~\ref{tab:scaling} shows that STAR's gains extend across model sizes, generations, and architectures.
The dense Qwen3.5-27B reaches an Avg6 of 81.10\%, exceeding Gemini 3.1 Pro by 3.06 percentage points.
The 4.67-point improvement on Qwen3-VL-32B-Instruct further demonstrates its applicability to checkpoints already tuned for instruction following.
Meanwhile, all three MoE students achieve consistent gains, with Qwen3.6-35B-A3B improving by 5.28 points to 80.89\%.
With only 3B active parameters, this model approaches the accuracy of the dense 27B model and surpasses some larger dense models, demonstrating high accuracy with substantially fewer active parameters.

\begin{figure}[!b]
    \centering
    \captionsetup{skip=2pt}
    \includegraphics[width=\columnwidth]{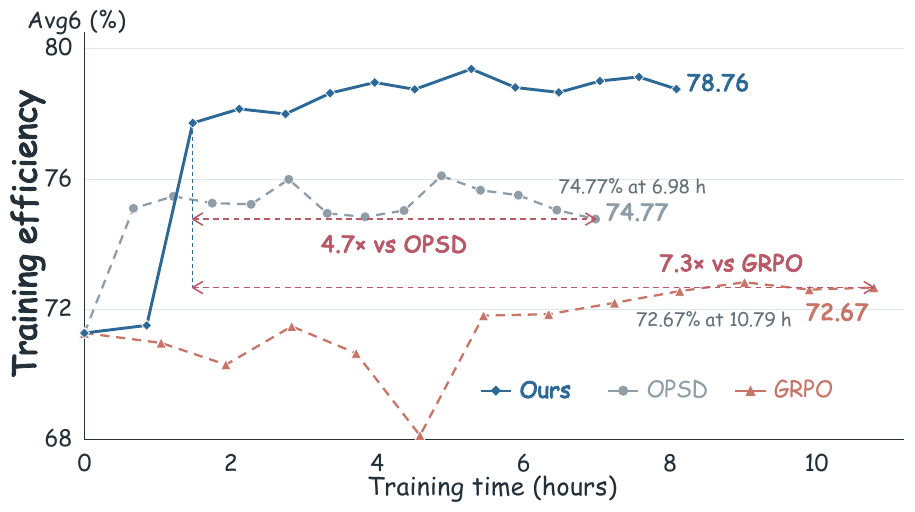}
    \caption{Avg6 versus measured training time, including training computation and checkpoint saving. Arrows compare the final OPSD and GRPO accuracies with the first measured STAR checkpoint exceeding each target.}
    \label{fig:training_efficiency}
\end{figure}

\subsection{Ablation Studies}
\label{sec:ablations}

\input{tab/component_ablation}

\par\noindent\textbf{Method components.}
Table~\ref{tab:component_ablation} examines privileged teacher supervision, counterfactual dual world rollouts, visual contrast reconstruction, and spatial attention modulation on Qwen3.5-4B.
We compare three target regimes: raw teacher targets without reconstruction at $\lambda=0$, visual contrast reconstruction with a fixed strength of $\lambda=1$, and adaptive reconstruction with attention modulated strength $\lambda_{w,t}$.
Each regime is evaluated using either rollouts from the original world alone or independent rollouts from both counterfactual worlds, with the base model serving as a reference.
Privileged crop supervision improves the base student even without target reconstruction, while introducing visual contrast provides further gains.
Counterfactual dual world rollouts improve Avg6 over the corresponding single world setting under all three target regimes.
However, their benefit remains limited when the contrast strength is fixed.
At $\lambda=1$, adding dual world rollouts lowers HR-Bench 4K accuracy from 87.13\% to 84.75\%, leaving an Avg6 gain of only 0.10 points.
This pattern is consistent with uniform contrast overcorrecting some tokens that are already well grounded across the more diverse trajectories.
Spatial attention modulation addresses this mismatch by adapting the correction strength to the student's attention state at each token.
It improves both sampling settings and increases the Avg6 benefit of dual world rollouts from 0.10 to 0.55 points.
Full STAR reaches 79.31\% Avg6 and recovers HR-Bench 4K to 87.50\%, the highest score among the ablated settings.
These results support the complementarity between \emph{exposing local fact changes through independent trajectories} and \emph{adapting visual correction to the generation states visited by the student}.

\input{tab/objective_teacher_ablation}

\par\addvspace{4pt}
\par\noindent\textbf{Distillation objective and teacher updates.}
Table~4 compares distillation objectives and teacher update strategies.
With a frozen teacher, FKL leads on ZoomBench and RKL on MME-RealWorld CN, while JS achieves the best scores on the remaining four benchmarks and the highest Avg6 of 79.31\%, exceeding FKL and RKL by 0.75 and 1.49 points, respectively.
Under JS, the frozen teacher also outperforms EMA with an update rate of 0.05 by 4.61 points across Avg6 and improves all six benchmarks, supporting its use for target reconstruction throughout student training in both worlds.

\subsection{Analysis}
\label{sec:analysis}

\noindent\textbf{Role of visual supervision.}
Under the diagnostic learning-rate schedule, the middle panel of Figure~\ref{fig:teaser} compares the raw teacher target at $\lambda=0$, external loss weighting, explicit attention loss, and our adaptive target reconstruction.
Across all evaluated checkpoints from step 20 to step 130, adaptive target reconstruction consistently achieves the highest Avg6 among these four settings.
Over the last ten logged updates, explicit attention loss reduces the mean spatial attention KL divergence to 0.70 compared with 2.75 under adaptive reconstruction, yet the two settings reach 76.07\% and 78.76\% Avg6 at step 130, respectively.
Forcing a lower internal attention divergence does not directly translate into higher answer accuracy; using spatial attention divergence to modulate the target distribution in vocabulary space more effectively converts local visual differences into supervision.

\par\addvspace{4pt}
\noindent\textbf{Training efficiency.}
Using the diagnostic learning-rate schedule, we measure elapsed time covering training computation and checkpoint saving across three 4B experiments, each generating 1,536 responses per step.
As shown in Figure~\ref{fig:training_efficiency}, STAR first surpasses the final OPSD accuracy of 74.77\% at a cumulative time of 1.48 hours, whereas OPSD requires 6.98 hours to reach that score, corresponding to a 4.7-fold speedup in reaching the target accuracy.
The corresponding speedup relative to GRPO is 7.3-fold.
By step 130, STAR, OPSD, and GRPO reach Avg6 scores of 78.76\%, 74.77\%, and 72.67\%, respectively, demonstrating the sample efficiency of dense visual target reconstruction over text-hint self-distillation and outcome-reward reinforcement learning.

\par
\begin{figure}[!htb]
    \centering
    \captionsetup{skip=2pt}
    \includegraphics[width=\columnwidth]{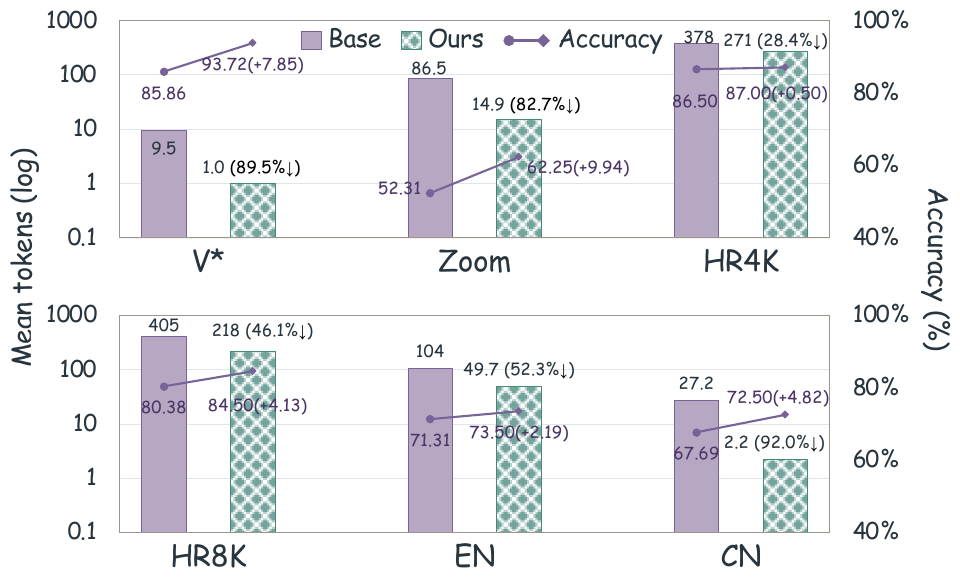}
    \caption{Mean output length and accuracy of Qwen3.5-9B. Bars compare Base and STAR output tokens on the left logarithmic axis; purple segments connect their accuracies on the right axis. EN and CN denote the MME-RealWorld English and Chinese subsets.}
    \label{fig:output_length}
\end{figure}

\addvspace{4pt}
\noindent\textbf{Output length.}
\label{sec:output_length}
We compute the average output length of responses from the Qwen3.5-9B checkpoints in Table~\ref{tab:main_results}, excluding input and EOS tokens. As shown in Figure~\ref{fig:output_length}, STAR reduces output tokens across all six benchmarks by 28.4\% to 92.0\% while improving accuracy by 0.50 to 9.94 points. On ZoomBench, for instance, the average response length drops from 86.5 to 14.9 tokens while accuracy increases by 9.94 points, indicating that stronger reliance on local visual evidence reduces redundant, speculative reasoning and enables accurate answers through more concise trajectories.

\input{sec/4_generalization}

\input{tab/generalization}

%% file: tab/main_results.tex
\begin{table*}[!t]
\centering
\begingroup
\definecolor{resultnegative}{RGB}{192,44,33}
\definecolor{resultours}{RGB}{220,235,250}
\caption{Main results on six fine-grained visual benchmarks (accuracy, \%). Colored annotations show percentage-point changes from the same-scale base (blue: increases; red: decreases). Bold marks the best score within each model group. Results marked with \textsuperscript{\textdagger} are taken from VAD~\cite{zhang2026vadattributingvisualevidence}; all other reported results are reproduced following the official methods and configurations under a unified evaluation protocol.}
\label{tab:main_results}
\fontsize{8.5}{9.5}\selectfont
\setlength{\tabcolsep}{1.8pt}
\renewcommand{\arraystretch}{0.96}
\newcommand{\resultentry}[2]{\makebox[22pt][r]{#1}\hspace{1pt}\makebox[20pt][r]{#2}}
\newcommand{\resultdelta}[2]{{\color{#1}\fontsize{6.5}{7}\selectfont #2}}
\newcommand{\resulticon}[1]{\makebox[8.5pt][c]{\raisebox{-1pt}{#1}}\hspace{2.5pt}}
\begin{tabularx}{\textwidth}{@{}l c *{7}{>{\raggedleft\arraybackslash}X}@{}}
\toprule
\textbf{Model} & \textbf{Param.} & \multicolumn{1}{c}{\textbf{V*}} & \multicolumn{1}{c}{\textbf{Zoom}} & \multicolumn{2}{c}{\textbf{HR-Bench}} & \multicolumn{2}{c}{\textbf{MME-RW}} & \multicolumn{1}{c}{\textbf{Avg6}} \\
\cmidrule(lr){5-6}\cmidrule(lr){7-8}
 & & & & \multicolumn{1}{c}{4K} & \multicolumn{1}{c}{8K} & \multicolumn{1}{c}{EN} & \multicolumn{1}{c}{CN} & \\
\midrule
\rowcolor{black!4}
\multicolumn{9}{c}{\textbf{Thinking-with-images agentic models}} \\
\resulticon{\includegraphics[width=7.5pt,height=7.5pt,keepaspectratio]{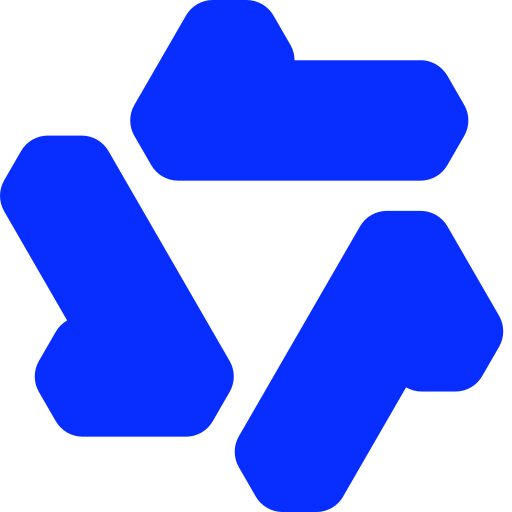}}DeepEyes~\cite{zheng2026deepeyes} & 7B & \resultentry{81.68}{} & \resultentry{46.15}{} & \resultentry{74.13}{} & \resultentry{69.63}{} & \resultentry{64.44}{} & \resultentry{64.91}{} & \resultentry{66.82}{} \\
\resulticon{\includegraphics[width=7.5pt,height=7.5pt,keepaspectratio]{fig/model-icons/qwen.png}}Thyme-RL~\cite{zhang2025thyme} & 7B & \resultentry{84.82}{} & \resultentry{46.86}{} & \resultentry{78.63}{} & \resultentry{72.88}{} & \resultentry{64.48}{} & \resultentry{63.65}{} & \resultentry{68.55}{} \\
\resulticon{\includegraphics[width=7.5pt,height=7.5pt,keepaspectratio]{fig/model-icons/qwen.png}}DeepEyesV2~\cite{hong2026deepeyesv2} & 7B & \resultentry{83.25}{} & \resultentry{46.75}{} & \resultentry{73.38}{} & \resultentry{71.75}{} & \resultentry{63.33}{} & \resultentry{63.29}{} & \resultentry{66.96}{} \\
\resulticon{\includegraphics[width=7.5pt,height=7.5pt,keepaspectratio]{fig/model-icons/qwen.png}}PixelReasoner~\cite{wang2025pixelreasoner} & 7B & \resultentry{78.53}{} & \resultentry{43.67}{} & \resultentry{67.50}{} & \resultentry{61.50}{} & \resultentry{63.58}{} & \resultentry{64.29}{} & \resultentry{63.18}{} \\
\resulticon{\includegraphics[width=7.5pt,height=7.5pt,keepaspectratio]{fig/model-icons/qwen.png}}VisReason~\cite{li2026visreason} & 7B & \resultentry{55.50}{} & \resultentry{34.91}{} & \resultentry{56.63}{} & \resultentry{52.00}{} & \resultentry{52.44}{} & \resultentry{47.03}{} & \resultentry{49.75}{} \\
\resulticon{\includegraphics[width=7.5pt,height=7.5pt,keepaspectratio]{fig/model-icons/qwen.png}}SenseNova-MARS~\cite{chng2025sensenovamars} & 8B & \resultentry{\textbf{92.15}}{} & \resultentry{\textbf{55.15}}{} & \resultentry{\textbf{84.13}}{} & \resultentry{\textbf{80.63}}{} & \resultentry{\textbf{69.63}}{} & \resultentry{\textbf{70.12}}{} & \resultentry{\textbf{75.30}}{} \\
\rowcolor{black!4}
\multicolumn{9}{c}{\textbf{Closed-source models}} \\
\resulticon{\includegraphics[width=7.5pt,height=7.5pt,keepaspectratio]{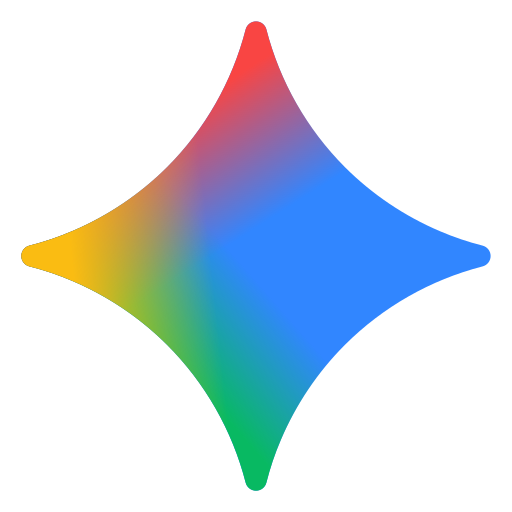}}Gemini 3 Flash~\cite{deepmind2025gemini3flash}\textsuperscript{\textdagger} & \resultlock & \resultentry{85.39}{} & \resultentry{61.49}{} & \resultentry{87.25}{} & \resultentry{\textbf{85.00}}{} & \resultentry{73.15}{} & \resultentry{71.66}{} & \resultentry{77.32}{} \\
\resulticon{\includegraphics[width=7.5pt,height=7.5pt,keepaspectratio]{fig/model-icons/gemini.png}}Gemini 3.1 Pro~\cite{deepmind2026gemini31pro}\textsuperscript{\textdagger} & \resultlock & \resultentry{\textbf{88.48}}{} & \resultentry{\textbf{62.01}}{} & \resultentry{\textbf{88.12}}{} & \resultentry{84.50}{} & \resultentry{73.32}{} & \resultentry{\textbf{71.81}}{} & \resultentry{\textbf{78.04}}{} \\
\resulticon{\includegraphics[width=7.5pt,height=7.5pt,keepaspectratio]{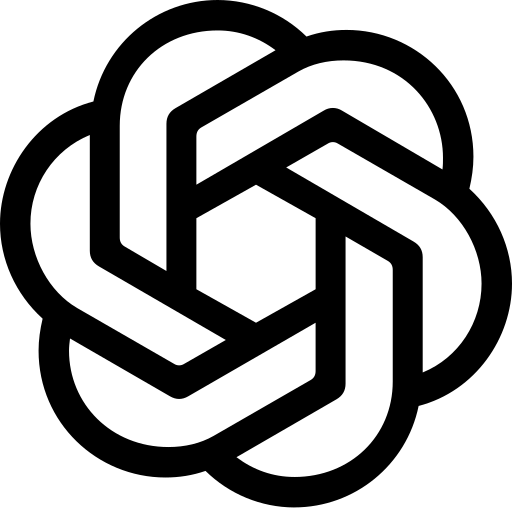}}GPT-5.2~\cite{openai2025gpt52}\textsuperscript{\textdagger} & \resultlock & \resultentry{84.59}{} & \resultentry{54.71}{} & \resultentry{84.64}{} & \resultentry{78.48}{} & \resultentry{71.82}{} & \resultentry{67.43}{} & \resultentry{73.61}{} \\
\resulticon{\includegraphics[width=7.5pt,height=7.5pt,keepaspectratio]{fig/model-icons/openai.png}}GPT-5.4~\cite{openai2026gpt54}\textsuperscript{\textdagger} & \resultlock & \resultentry{85.83}{} & \resultentry{57.94}{} & \resultentry{86.15}{} & \resultentry{78.10}{} & \resultentry{\textbf{74.16}}{} & \resultentry{70.50}{} & \resultentry{75.45}{} \\
\rowcolor{black!4}
\multicolumn{9}{c}{\textbf{Large open-source models}} \\
\resulticon{\includegraphics[width=7.5pt,height=7.5pt,keepaspectratio]{fig/model-icons/qwen.png}}Qwen3.5-35B-A3B Base~\cite{qwen2026qwen35} & 35B & \resultentry{83.77}{} & \resultentry{52.66}{} & \resultentry{87.38}{} & \resultentry{84.38}{} & \resultentry{72.20}{} & \resultentry{68.38}{} & \resultentry{74.79}{} \\
\resulticon{\includegraphics[width=7.5pt,height=7.5pt,keepaspectratio]{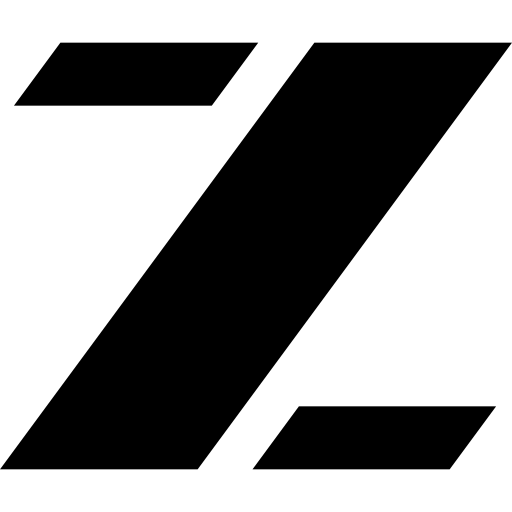}}GLM-4.6V~\cite{zai2025glm46v} & 106B & \resultentry{81.15}{} & \resultentry{47.57}{} & \resultentry{79.25}{} & \resultentry{75.63}{} & \resultentry{67.72}{} & \resultentry{65.19}{} & \resultentry{69.42}{} \\
\resulticon{\includegraphics[width=7.5pt,height=7.5pt,keepaspectratio]{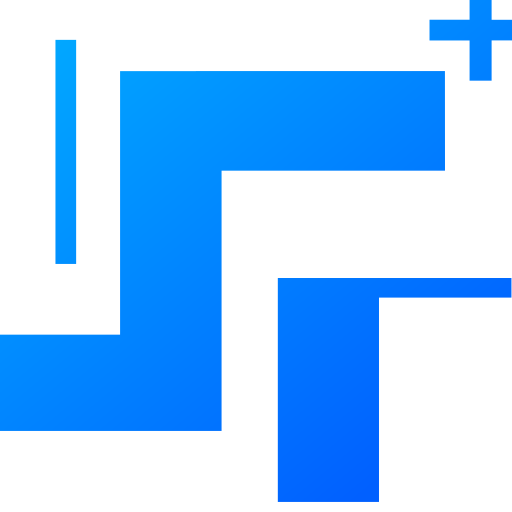}}Step-3.7-Flash~\cite{stepfun2026step37flash} & 198B & \resultentry{\textbf{89.01}}{} & \resultentry{51.48}{} & \resultentry{81.50}{} & \resultentry{74.50}{} & \resultentry{67.04}{} & \resultentry{62.89}{} & \resultentry{71.07}{} \\
\resulticon{\includegraphics[width=7.5pt,height=7.5pt,keepaspectratio]{fig/model-icons/qwen.png}}Qwen3-VL-235B-Instruct~\cite{qwen2025qwen3vl} & 235B & \resultentry{\textbf{89.01}}{} & \resultentry{54.67}{} & \resultentry{84.38}{} & \resultentry{82.25}{} & \resultentry{71.30}{} & \resultentry{\textbf{68.87}}{} & \resultentry{75.08}{} \\
\resulticon{\includegraphics[width=7.5pt,height=7.5pt,keepaspectratio]{fig/model-icons/zai.png}}GLM-5.3-Flash~\cite{zai2026glm53flash} & 320B & \resultentry{79.06}{} & \resultentry{50.53}{} & \resultentry{71.38}{} & \resultentry{62.62}{} & \resultentry{66.57}{} & \resultentry{60.06}{} & \resultentry{65.04}{} \\
\resulticon{\includegraphics[width=7.5pt,height=7.5pt,keepaspectratio]{fig/model-icons/qwen.png}}Qwen3.5-397B-A17B~\cite{qwen2026qwen35} & 397B & \resultentry{85.34}{} & \resultentry{\textbf{57.99}}{} & \resultentry{\textbf{88.13}}{} & \resultentry{\textbf{85.50}}{} & \resultentry{\textbf{74.07}}{} & \resultentry{67.97}{} & \resultentry{\textbf{76.50}}{} \\
\resulticon{\includegraphics[width=7.5pt,height=7.5pt,keepaspectratio]{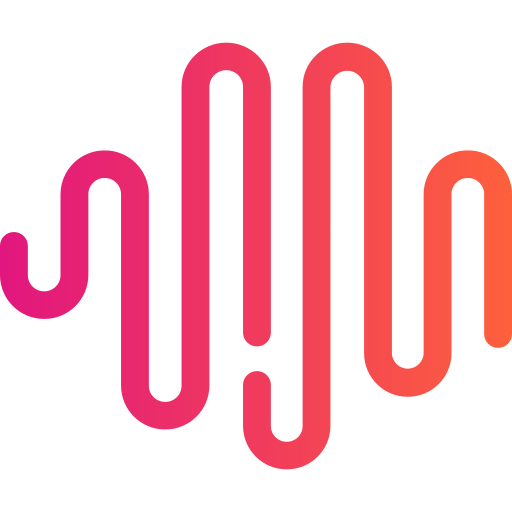}}MiniMax-M3~\cite{minimax2026m3} & 428B & \resultentry{59.16}{} & \resultentry{46.27}{} & \resultentry{66.00}{} & \resultentry{58.88}{} & \resultentry{55.72}{} & \resultentry{45.29}{} & \resultentry{55.22}{} \\
\resulticon{\includegraphics[width=7.5pt,height=7.5pt,keepaspectratio]{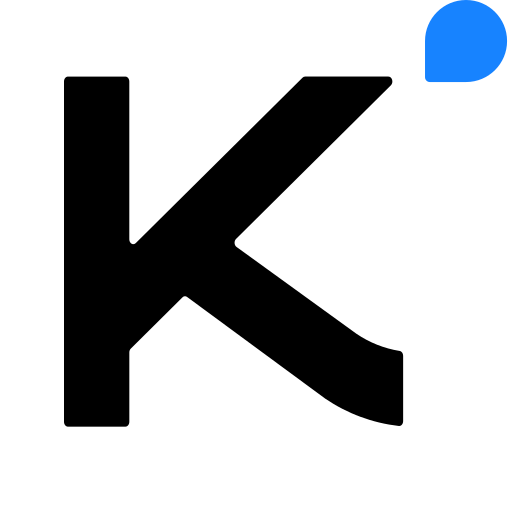}}Kimi-K2.6~\cite{moonshot2026kimik26} & 1T & \resultentry{84.29}{} & \resultentry{53.25}{} & \resultentry{81.38}{} & \resultentry{75.25}{} & \resultentry{66.87}{} & \resultentry{66.99}{} & \resultentry{71.34}{} \\
\rowcolor{black!4}
\multicolumn{9}{c}{\textbf{Qwen3.5-based models: 4B}} \\
\resulticon{\includegraphics[width=7.5pt,height=7.5pt,keepaspectratio]{fig/model-icons/qwen.png}}Qwen3.5 Base~\cite{qwen2026qwen35} & 4B & \resultentry{81.68}{} & \resultentry{49.35}{} & \resultentry{86.25}{} & \resultentry{82.75}{} & \resultentry{63.81}{} & \resultentry{63.77}{} & \resultentry{71.27}{} \\
\resulticon{\includegraphics[width=7.5pt,height=7.5pt,keepaspectratio]{fig/model-icons/qwen.png}}GRPO~\cite{shao2024deepseekmath} & 4B & \resultentry{80.10}{\resultdelta{resultnegative}{-1.58}} & \resultentry{57.99}{\resultdelta{takeawayblue}{+8.64}} & \resultentry{85.25}{\resultdelta{resultnegative}{-1.00}} & \resultentry{82.25}{\resultdelta{resultnegative}{-0.50}} & \resultentry{73.13}{\resultdelta{takeawayblue}{+9.32}} & \resultentry{71.00}{\resultdelta{takeawayblue}{+7.23}} & \resultentry{74.95}{\resultdelta{takeawayblue}{+3.68}} \\
\resulticon{\includegraphics[width=7.5pt,height=7.5pt,keepaspectratio]{fig/model-icons/qwen.png}}OPSD~\cite{zhao2026selfdistilled} & 4B & \resultentry{83.77}{\resultdelta{takeawayblue}{+2.09}} & \resultentry{53.14}{\resultdelta{takeawayblue}{+3.79}} & \resultentry{85.25}{\resultdelta{resultnegative}{-1.00}} & \resultentry{77.88}{\resultdelta{resultnegative}{-4.88}} & \resultentry{75.05}{\resultdelta{takeawayblue}{+11.24}} & \resultentry{\textbf{74.45}}{\resultdelta{takeawayblue}{+10.68}} & \resultentry{74.92}{\resultdelta{takeawayblue}{+3.65}} \\
\resulticon{\includegraphics[width=7.5pt,height=7.5pt,keepaspectratio]{fig/model-icons/qwen.png}}Vision-OPD~\cite{yuan2026visionopdlearningfinedetails} & 4B & \resultentry{87.96}{\resultdelta{takeawayblue}{+6.28}} & \resultentry{60.24}{\resultdelta{takeawayblue}{+10.89}} & \resultentry{81.75}{\resultdelta{resultnegative}{-4.50}} & \resultentry{80.12}{\resultdelta{resultnegative}{-2.63}} & \resultentry{75.01}{\resultdelta{takeawayblue}{+11.20}} & \resultentry{70.51}{\resultdelta{takeawayblue}{+6.74}} & \resultentry{75.93}{\resultdelta{takeawayblue}{+4.66}} \\
\resulticon{\includegraphics[width=7.5pt,height=7.5pt,keepaspectratio]{fig/model-icons/qwen.png}}OPD-V~\cite{aniri2026opdvvisualonpolicyselfdistillation} & 4B & \resultentry{90.58}{\resultdelta{takeawayblue}{+8.90}} & \resultentry{59.53}{\resultdelta{takeawayblue}{+10.18}} & \resultentry{85.25}{\resultdelta{resultnegative}{-1.00}} & \resultentry{80.75}{\resultdelta{resultnegative}{-2.00}} & \resultentry{75.62}{\resultdelta{takeawayblue}{+11.81}} & \resultentry{72.01}{\resultdelta{takeawayblue}{+8.24}} & \resultentry{77.29}{\resultdelta{takeawayblue}{+6.02}} \\
\resulticon{\includegraphics[width=7.5pt,height=7.5pt,keepaspectratio]{fig/model-icons/qwen.png}}VA-OPD~\cite{liu2026visualadvantageonpolicydistillationvisionlanguage} & 4B & \resultentry{90.58}{\resultdelta{takeawayblue}{+8.90}} & \resultentry{58.58}{\resultdelta{takeawayblue}{+9.23}} & \resultentry{83.88}{\resultdelta{resultnegative}{-2.37}} & \resultentry{80.12}{\resultdelta{resultnegative}{-2.63}} & \resultentry{75.95}{\resultdelta{takeawayblue}{+12.14}} & \resultentry{70.86}{\resultdelta{takeawayblue}{+7.09}} & \resultentry{76.66}{\resultdelta{takeawayblue}{+5.39}} \\
\resulticon{\includegraphics[width=7.5pt,height=7.5pt,keepaspectratio]{fig/model-icons/qwen.png}}V-Zero~\cite{sun2026vzeroanswerlabelfreeonpolicydistillation} & 4B & \resultentry{89.01}{\resultdelta{takeawayblue}{+7.33}} & \resultentry{59.64}{\resultdelta{takeawayblue}{+10.29}} & \resultentry{86.38}{\resultdelta{takeawayblue}{+0.13}} & \resultentry{81.75}{\resultdelta{resultnegative}{-1.00}} & \resultentry{75.48}{\resultdelta{takeawayblue}{+11.67}} & \resultentry{70.53}{\resultdelta{takeawayblue}{+6.76}} & \resultentry{77.13}{\resultdelta{takeawayblue}{+5.86}} \\
\resulticon{\includegraphics[width=7.5pt,height=7.5pt,keepaspectratio]{fig/model-icons/qwen.png}}CV-OPD~\cite{wang2026cvopdcontrastivevisiononpolicy} & 4B & \resultentry{89.53}{\resultdelta{takeawayblue}{+7.85}} & \resultentry{\textbf{61.89}}{\resultdelta{takeawayblue}{+12.54}} & \resultentry{83.88}{\resultdelta{resultnegative}{-2.38}} & \resultentry{80.13}{\resultdelta{resultnegative}{-2.63}} & \resultentry{75.55}{\resultdelta{takeawayblue}{+11.74}} & \resultentry{70.63}{\resultdelta{takeawayblue}{+6.86}} & \resultentry{76.93}{\resultdelta{takeawayblue}{+5.66}} \\
\resulticon{\includegraphics[width=7.5pt,height=7.5pt,keepaspectratio]{fig/model-icons/qwen.png}}VAD~\cite{zhang2026vadattributingvisualevidence} & 4B & \resultentry{93.72}{\resultdelta{takeawayblue}{+12.04}} & \resultentry{60.83}{\resultdelta{takeawayblue}{+11.48}} & \resultentry{86.00}{\resultdelta{resultnegative}{-0.25}} & \resultentry{83.00}{\resultdelta{takeawayblue}{+0.25}} & \resultentry{76.68}{\resultdelta{takeawayblue}{+12.87}} & \resultentry{71.30}{\resultdelta{takeawayblue}{+7.53}} & \resultentry{78.59}{\resultdelta{takeawayblue}{+7.32}} \\
\rowcolor{resultours}
\resulticon{\includegraphics[width=7.5pt,height=7.5pt,keepaspectratio]{fig/model-icons/qwen.png}}\textcolor{takeawayblue}{\textbf{Ours}} & \textcolor{takeawayblue}{4B} & \textcolor{takeawayblue}{\resultentry{\textbf{94.76}}{\resultdelta{takeawayblue}{+13.08}}} & \textcolor{takeawayblue}{\resultentry{60.95}{\resultdelta{takeawayblue}{+11.60}}} & \textcolor{takeawayblue}{\resultentry{\textbf{87.50}}{\resultdelta{takeawayblue}{+1.25}}} & \textcolor{takeawayblue}{\resultentry{\textbf{84.38}}{\resultdelta{takeawayblue}{+1.63}}} & \textcolor{takeawayblue}{\resultentry{\textbf{77.76}}{\resultdelta{takeawayblue}{+13.95}}} & \textcolor{takeawayblue}{\resultentry{70.51}{\resultdelta{takeawayblue}{+6.74}}} & \textcolor{takeawayblue}{\resultentry{\textbf{79.31}}{\resultdelta{takeawayblue}{+8.04}}} \\
\rowcolor{black!4}
\multicolumn{9}{c}{\textbf{Qwen3.5-based models: 9B}} \\
\resulticon{\includegraphics[width=7.5pt,height=7.5pt,keepaspectratio]{fig/model-icons/qwen.png}}Qwen3.5 Base~\cite{qwen2026qwen35} & 9B & \resultentry{85.86}{} & \resultentry{52.31}{} & \resultentry{86.50}{} & \resultentry{80.38}{} & \resultentry{71.31}{} & \resultentry{67.69}{} & \resultentry{74.01}{} \\
\resulticon{\includegraphics[width=7.5pt,height=7.5pt,keepaspectratio]{fig/model-icons/qwen.png}}GRPO~\cite{shao2024deepseekmath} & 9B & \resultentry{87.43}{\resultdelta{takeawayblue}{+1.57}} & \resultentry{58.11}{\resultdelta{takeawayblue}{+5.80}} & \resultentry{86.00}{\resultdelta{resultnegative}{-0.50}} & \resultentry{84.63}{\resultdelta{takeawayblue}{+4.25}} & \resultentry{71.85}{\resultdelta{takeawayblue}{+0.54}} & \resultentry{70.10}{\resultdelta{takeawayblue}{+2.42}} & \resultentry{76.35}{\resultdelta{takeawayblue}{+2.35}} \\
\resulticon{\includegraphics[width=7.5pt,height=7.5pt,keepaspectratio]{fig/model-icons/qwen.png}}OPSD~\cite{zhao2026selfdistilled} & 9B & \resultentry{89.53}{\resultdelta{takeawayblue}{+3.67}} & \resultentry{55.15}{\resultdelta{takeawayblue}{+2.84}} & \resultentry{84.00}{\resultdelta{resultnegative}{-2.50}} & \resultentry{81.00}{\resultdelta{takeawayblue}{+0.62}} & \resultentry{75.01}{\resultdelta{takeawayblue}{+3.70}} & \resultentry{72.45}{\resultdelta{takeawayblue}{+4.76}} & \resultentry{76.19}{\resultdelta{takeawayblue}{+2.18}} \\
\resulticon{\includegraphics[width=7.5pt,height=7.5pt,keepaspectratio]{fig/model-icons/qwen.png}}Vision-OPD~\cite{yuan2026visionopdlearningfinedetails} & 9B & \resultentry{84.82}{\resultdelta{resultnegative}{-1.05}} & \resultentry{\textbf{63.67}}{\resultdelta{takeawayblue}{+11.36}} & \resultentry{84.88}{\resultdelta{resultnegative}{-1.63}} & \resultentry{82.00}{\resultdelta{takeawayblue}{+1.63}} & \resultentry{71.68}{\resultdelta{takeawayblue}{+0.37}} & \resultentry{69.55}{\resultdelta{takeawayblue}{+1.86}} & \resultentry{76.10}{\resultdelta{takeawayblue}{+2.09}} \\
\resulticon{\includegraphics[width=7.5pt,height=7.5pt,keepaspectratio]{fig/model-icons/qwen.png}}OPD-V~\cite{aniri2026opdvvisualonpolicyselfdistillation} & 9B & \resultentry{85.34}{\resultdelta{resultnegative}{-0.52}} & \resultentry{60.59}{\resultdelta{takeawayblue}{+8.28}} & \resultentry{85.63}{\resultdelta{resultnegative}{-0.88}} & \resultentry{82.75}{\resultdelta{takeawayblue}{+2.38}} & \resultentry{72.87}{\resultdelta{takeawayblue}{+1.56}} & \resultentry{70.98}{\resultdelta{takeawayblue}{+3.30}} & \resultentry{76.36}{\resultdelta{takeawayblue}{+2.35}} \\
\resulticon{\includegraphics[width=7.5pt,height=7.5pt,keepaspectratio]{fig/model-icons/qwen.png}}VA-OPD~\cite{liu2026visualadvantageonpolicydistillationvisionlanguage} & 9B & \resultentry{89.01}{\resultdelta{takeawayblue}{+3.14}} & \resultentry{61.42}{\resultdelta{takeawayblue}{+9.11}} & \resultentry{\textbf{88.88}}{\resultdelta{takeawayblue}{+2.38}} & \resultentry{85.00}{\resultdelta{takeawayblue}{+4.63}} & \resultentry{74.92}{\resultdelta{takeawayblue}{+3.62}} & \resultentry{70.46}{\resultdelta{takeawayblue}{+2.77}} & \resultentry{78.28}{\resultdelta{takeawayblue}{+4.27}} \\
\resulticon{\includegraphics[width=7.5pt,height=7.5pt,keepaspectratio]{fig/model-icons/qwen.png}}V-Zero~\cite{sun2026vzeroanswerlabelfreeonpolicydistillation} & 9B & \resultentry{86.39}{\resultdelta{takeawayblue}{+0.52}} & \resultentry{62.37}{\resultdelta{takeawayblue}{+10.06}} & \resultentry{87.63}{\resultdelta{takeawayblue}{+1.13}} & \resultentry{85.50}{\resultdelta{takeawayblue}{+5.13}} & \resultentry{75.36}{\resultdelta{takeawayblue}{+4.05}} & \resultentry{68.02}{\resultdelta{takeawayblue}{+0.34}} & \resultentry{77.54}{\resultdelta{takeawayblue}{+3.54}} \\
\resulticon{\includegraphics[width=7.5pt,height=7.5pt,keepaspectratio]{fig/model-icons/qwen.png}}CV-OPD~\cite{wang2026cvopdcontrastivevisiononpolicy} & 9B & \resultentry{86.91}{\resultdelta{takeawayblue}{+1.05}} & \resultentry{62.96}{\resultdelta{takeawayblue}{+10.65}} & \resultentry{85.88}{\resultdelta{resultnegative}{-0.63}} & \resultentry{83.38}{\resultdelta{takeawayblue}{+3.00}} & \resultentry{72.69}{\resultdelta{takeawayblue}{+1.39}} & \resultentry{70.95}{\resultdelta{takeawayblue}{+3.26}} & \resultentry{77.13}{\resultdelta{takeawayblue}{+3.12}} \\
\resulticon{\includegraphics[width=7.5pt,height=7.5pt,keepaspectratio]{fig/model-icons/qwen.png}}VAD~\cite{zhang2026vadattributingvisualevidence} & 9B & \resultentry{87.96}{\resultdelta{takeawayblue}{+2.10}} & \resultentry{62.13}{\resultdelta{takeawayblue}{+9.82}} & \resultentry{87.25}{\resultdelta{takeawayblue}{+0.75}} & \resultentry{\textbf{87.63}}{\resultdelta{takeawayblue}{+7.25}} & \resultentry{\textbf{75.82}}{\resultdelta{takeawayblue}{+4.51}} & \resultentry{70.07}{\resultdelta{takeawayblue}{+2.38}} & \resultentry{78.48}{\resultdelta{takeawayblue}{+4.47}} \\
\rowcolor{resultours}
\resulticon{\includegraphics[width=7.5pt,height=7.5pt,keepaspectratio]{fig/model-icons/qwen.png}}\textcolor{takeawayblue}{\textbf{Ours}} & \textcolor{takeawayblue}{9B} & \textcolor{takeawayblue}{\resultentry{\textbf{93.72}}{\resultdelta{takeawayblue}{+7.85}}} & \textcolor{takeawayblue}{\resultentry{62.25}{\resultdelta{takeawayblue}{+9.94}}} & \textcolor{takeawayblue}{\resultentry{87.00}{\resultdelta{takeawayblue}{+0.50}}} & \textcolor{takeawayblue}{\resultentry{84.50}{\resultdelta{takeawayblue}{+4.13}}} & \textcolor{takeawayblue}{\resultentry{73.50}{\resultdelta{takeawayblue}{+2.19}}} & \textcolor{takeawayblue}{\resultentry{\textbf{72.50}}{\resultdelta{takeawayblue}{+4.82}}} & \textcolor{takeawayblue}{\resultentry{\textbf{78.91}}{\resultdelta{takeawayblue}{+4.90}}} \\
\bottomrule
\end{tabularx}
\endgroup
\end{table*}

%% file: tab/scaling.tex
\begin{table}[t]
\centering
\caption{Models scale and architecture. Avg6 accuracy (\%) and gains in percentage points.}
\label{tab:scaling}
\begingroup
\fontsize{9}{10.5}\selectfont
\setlength{\tabcolsep}{3pt}
\renewcommand{\arraystretch}{1.04}
\begin{tabular*}{\columnwidth}{@{\extracolsep{\fill}}lrrr@{}}
\toprule
\textbf{Models} & \textbf{Base} & \textbf{STAR} & \textbf{$\Delta$Avg6} \\
\midrule
Qwen3.5-27B & 76.28 & \textcolor{takeawayblue}{\textbf{81.10}} & \textcolor{takeawayblue}{+4.82} \\
Qwen3.8-27B & 75.01 & \textcolor{takeawayblue}{76.67} & \textcolor{takeawayblue}{+1.66} \\
Qwen3-VL-32B-Instruct & 72.78 & \textcolor{takeawayblue}{77.45} & \textcolor{takeawayblue}{+4.67} \\
\addlinespace[2pt]
Qwen3.5-35B-A3B & 74.79 & \textcolor{takeawayblue}{79.27} & \textcolor{takeawayblue}{+4.48} \\
Qwen3.6-35B-A3B & 75.61 & \textcolor{takeawayblue}{80.89} & \textcolor{takeawayblue}{+5.28} \\
Qwen3.5-122B-A10B & 74.85 & \textcolor{takeawayblue}{80.82} & \textcolor{takeawayblue}{+5.97} \\
\bottomrule
\end{tabular*}
\endgroup
\end{table}

%% file: tab/component_ablation.tex
\begin{table*}[t]
\centering
\caption{Component ablations on Qwen3.5-4B (accuracy, \%). Single- and dual-world settings are paired under each reconstruction rule; dual-world rollouts are sampled independently. Bold marks the best result in each column.}
\label{tab:component_ablation}
\begingroup
\definecolor{ablationours}{RGB}{220,235,250}
\fontsize{8.5}{10}\selectfont
\setlength{\tabcolsep}{1.8pt}
\renewcommand{\arraystretch}{1.05}
\begin{tabularx}{\linewidth}{@{}l c c c c *{7}{>{\raggedleft\arraybackslash}X}@{}}
\toprule
\textbf{Setting} & \shortstack{\textbf{Privileged}\\\textbf{teacher}} & \shortstack{\textbf{Dual-world}\\\textbf{Rollouts}} & \shortstack{\textbf{Visual}\\\textbf{Contrast}} & \shortstack{\textbf{Attention}\\\textbf{Modulation}} & \textbf{V*} & \textbf{Zoom} & \textbf{HR4K} & \textbf{HR8K} & \shortstack{\textbf{MME}\\\textbf{EN}} & \shortstack{\textbf{MME}\\\textbf{CN}} & \textbf{Avg6} \\
\midrule
Base & \textemdash & \textemdash & \textemdash & \textemdash & 81.68 & 49.35 & 86.25 & 82.75 & 63.81 & 63.77 & 71.27 \\
\addlinespace[1.5pt]
Single-world ($\lambda=0$) & $\checkmark$ & \textemdash & \textemdash & \textemdash & 87.96 & 57.87 & 83.88 & 82.88 & 74.65 & 67.79 & 75.84 \\
Dual-world ($\lambda=0$) & $\checkmark$ & $\checkmark$ & \textemdash & \textemdash & 91.10 & 58.70 & 84.63 & 80.88 & 75.41 & 68.16 & 76.48 \\
\addlinespace[1.5pt]
Single-world (fixed $\lambda=1$) & $\checkmark$ & \textemdash & $\checkmark$ & \textemdash & 92.67 & 60.12 & 87.13 & 82.50 & 77.05 & 70.69 & 78.36 \\
Dual-world (fixed $\lambda=1$) & $\checkmark$ & $\checkmark$ & $\checkmark$ & \textemdash & 93.72 & 60.12 & 84.75 & 84.13 & 77.12 & \textbf{70.95} & 78.46 \\
\addlinespace[1.5pt]
Single-world (adaptive) & $\checkmark$ & \textemdash & $\checkmark$ & $\checkmark$ & 93.72 & \textbf{61.18} & 87.00 & 82.88 & 76.90 & 70.88 & 78.76 \\
\rowcolor{ablationours}
\textcolor{takeawayblue}{\textbf{Full method}} & \textcolor{takeawayblue}{$\checkmark$} & \textcolor{takeawayblue}{$\checkmark$} & \textcolor{takeawayblue}{$\checkmark$} & \textcolor{takeawayblue}{$\checkmark$} & \textcolor{takeawayblue}{\textbf{94.76}} & \textcolor{takeawayblue}{60.95} & \textcolor{takeawayblue}{\textbf{87.50}} & \textcolor{takeawayblue}{\textbf{84.38}} & \textcolor{takeawayblue}{\textbf{77.76}} & \textcolor{takeawayblue}{70.51} & \textcolor{takeawayblue}{\textbf{79.31}} \\
\bottomrule
\end{tabularx}
\endgroup
\end{table*}

%% file: tab/objective_teacher_ablation.tex
\begin{table}[!htb]
\centering
\caption{Distillation objectives and teacher updates on Qwen3.5-4B (accuracy, \%). EMA uses an update rate of 0.05. Bold marks the best result in each column.}
\label{tab:objective_teacher_ablation}
\begingroup
\definecolor{ablationours}{RGB}{220,235,250}
\fontsize{8.2}{9.5}\selectfont
\setlength{\tabcolsep}{1.2pt}
\renewcommand{\arraystretch}{1.05}
\begin{tabularx}{\linewidth}{@{}l l *{7}{>{\raggedleft\arraybackslash}X}@{}}
\toprule
\textbf{Loss} & \textbf{Teacher} & \multicolumn{1}{c}{\textbf{V*}} & \multicolumn{1}{c}{\textbf{Zoom}} & \multicolumn{2}{c}{\textbf{HR-Bench}} & \multicolumn{2}{c}{\textbf{MME-RW}} & \multicolumn{1}{c}{\textbf{Avg6}} \\
\cmidrule(lr){5-6}\cmidrule(lr){7-8}
 & & & & \multicolumn{1}{c}{4K} & \multicolumn{1}{c}{8K} & \multicolumn{1}{c}{EN} & \multicolumn{1}{c}{CN} & \\
\midrule
FKL & Frozen & 91.10 & \textbf{61.42} & 87.13 & \textbf{84.38} & 76.35 & 71.00 & 78.56 \\
RKL & Frozen & 91.62 & 59.05 & 86.63 & 81.13 & 76.19 & \textbf{72.28} & 77.82 \\
\addlinespace[2pt]
 & EMA & 86.91 & 60.36 & 82.13 & 78.63 & 70.90 & 69.27 & 74.70 \\
\rowcolor{ablationours}
\textcolor{takeawayblue}{\raisebox{5pt}[0pt][0pt]{\textbf{JS}}} & \textcolor{takeawayblue}{\textbf{Frozen}} & \textcolor{takeawayblue}{\textbf{94.76}} & \textcolor{takeawayblue}{60.95} & \textcolor{takeawayblue}{\textbf{87.50}} & \textcolor{takeawayblue}{\textbf{84.38}} & \textcolor{takeawayblue}{\textbf{77.76}} & \textcolor{takeawayblue}{70.51} & \textcolor{takeawayblue}{\textbf{79.31}} \\
\bottomrule
\end{tabularx}
\endgroup
\end{table}

%% file: sec/4_generalization.tex
\par\addvspace{4pt}
\noindent\textbf{Generalization.}
\label{sec:generalization}
Table~\ref{tab:generalization} evaluates out-of-domain generalization across four additional benchmarks. STAR improves over the base model on every benchmark across all three scales: accuracy on DynaMath increases by 22.36, 7.01, and 8.60 percentage points at the 4B, 9B, and 27B scales, respectively, while MMVP improves by 3.33 to 6.00 percentage points, CV-Bench improves by 0.72 to 1.64 percentage points, and the 27B model gains 6.70 percentage points on LogicVista. These results confirm that enhancing local perception via counterfactual dual worlds and attention guided reconstruction transfers to broader mathematical, discriminative, and logical visual reasoning on unseen benchmarks.

%% file: tab/generalization.tex
\begin{table}[t]
\centering
\caption{Generalization on four additional benchmarks. Scores are accuracy (\%).}
\label{tab:generalization}
\begingroup
\definecolor{generalizationours}{RGB}{220,235,250}
\fontsize{8.5}{10}\selectfont
\setlength{\tabcolsep}{1.4pt}
\renewcommand{\arraystretch}{1.08}
\newcommand{\oodscore}[2]{\textcolor{takeawayblue}{\makebox[22pt][r]{\textbf{#1}}\hspace{1pt}{\fontsize{6.4}{7}\selectfont+#2}}}
\begin{tabularx}{\linewidth}{@{}l *{4}{>{\centering\arraybackslash}X}@{}}
\toprule
\textbf{Model} & \textbf{DynaMath} & \textbf{MMVP} & \textbf{LogicVista} & \textbf{CV-Bench} \\
\midrule
4B Base & 55.61 & 60.00 & 64.96 & 86.15 \\
\rowcolor{generalizationours}
\textcolor{takeawayblue}{\textbf{4B Ours}} & \oodscore{77.96}{22.36} & \oodscore{66.00}{6.00} & \oodscore{65.18}{0.22} & \oodscore{86.87}{0.72} \\
\addlinespace[2pt]
9B Base & 70.20 & 66.67 & 65.18 & 86.34 \\
\rowcolor{generalizationours}
\textcolor{takeawayblue}{\textbf{9B Ours}} & \oodscore{77.21}{7.01} & \oodscore{72.67}{6.00} & \oodscore{66.74}{1.56} & \oodscore{87.42}{1.07} \\
\addlinespace[2pt]
27B Base & 70.36 & 71.33 & 69.42 & 87.80 \\
\rowcolor{generalizationours}
\textcolor{takeawayblue}{\textbf{27B Ours}} & \oodscore{78.96}{8.60} & \oodscore{74.67}{3.33} & \oodscore{76.12}{6.70} & \oodscore{89.44}{1.64} \\
\bottomrule
\end{tabularx}
\endgroup
\end{table}

%% file: sec/5_conclusion.tex
\section{Conclusion}
\label{sec:conclusion}

We presented \textbf{STAR}, a framework for visual on policy self distillation that integrates independent rollouts in paired counterfactual worlds with target reconstruction guided by spatial attention.
Sampling trajectories independently in worlds with altered local facts exposes contextual shortcuts while keeping each prefix consistent with its image, whereas spatial attention divergence between the student and teacher modulates the teacher contrast between real evidence crops and crops filled with the mean color to adapt target distributions at each token.
Extensive experiments across dense and MoE architectures demonstrate that STAR consistently improves perception of fine visual details, training efficiency, and generalization beyond the training domain.

%% file: sec/llm_usage.tex
\section*{LLM Usage Statement}
\noindent We used large language models to improve the clarity and fluency of the manuscript and assist with figure preparation. The scientific ideas, methods, experimental design, and conclusions were developed by the authors. The authors reviewed and verified all assisted content and take full responsibility for the final manuscript.